\documentclass{article}
\usepackage[utf8]{inputenc}
\usepackage{authblk}
\usepackage{setspace}
\usepackage[margin=1.25in]{geometry}
\usepackage{graphicx}
\graphicspath{ {./figures/} }
\usepackage{widetext}
\usepackage{amsmath}
\usepackage{algorithm}
\usepackage{algorithmic}
\usepackage{lineno}
\usepackage{color}
\usepackage[caption=false,font=normalsize,labelfont=sf,textfont=sf]{subfig}
\usepackage[numbers,sort&compress]{natbib}

\title{Quantitative Representation of Scenario Difficulty for Autonomous Driving Based on Adversarial Policy Search}

\author[1]{Shuo Yang}
\author[1]{Caojun Wang}
\author[3]{Yuanjian Zhang}
\author[4]{Yuming Yin}
\author[1,2*]{Yanjun Huang}
\author[5]{Shengbo Eben Li}
\author[6]{Hong Chen}

\affil[1]{School of Automotive studies, Tongji University, Shanghai, 201804, China.}
\affil[2]{Frontiers Science Center for Intelligent Autonomous Systems, Shanghai, 200120, China.}
\affil[3]{Department of Aeronautical and Automotive Engineering, Loughborough University, Leicestershire LE11 3TU, UK.}
\affil[4]{School of Mechanical Engineering, Zhejiang University of Technology, Hangzhou, 310014, China.}
\affil[5]{School of Vehicle and Mobility, Tsinghua University, Beijing, 100084, China.}
\affil[6]{College of Electronics and Information Engineering, Tongji University, Shanghai 201804, China.}
\affil[*]{Address correspondence to: yanjun\_huang@tongji.edu.cn}

\date{}

\begin{document}

\maketitle

\begin{abstract}
Adversarial scenario generation is crucial for autonomous driving testing because it can efficiently simulate various challenge and complex traffic conditions. However, it is difficult to control current existing methods to generate desired scenarios, such as the ones with different conflict levels. Therefore, this paper proposes a data-driven quantitative method to represent scenario difficulty. Compared with rule-based discrete scenario difficulty representation method, the proposed algorithm can achieve continuous difficulty representation. Specifically, the environment agent is introduced, and a reinforcement learning method combined with mechanism knowledge is constructed for policy search to obtain an agent with adversarial behavior. The model parameters of the environment agent at different stages in the training process are extracted to construct a policy group, and then the agents with different adversarial intensity are obtained, which are used to realize data generation in different difficulty scenarios through the simulation environment. Finally, a data-driven scenario difficulty quantitative representation model is constructed, which is used to output the environment agent policy under different difficulties. The result analysis shows that the proposed algorithm can generate reasonable and interpretable scenarios with high discrimination, and can provide quantifiable difficulty representation without any expert logic rule design. The video link is https://www.youtube.com/watch?v=GceGdqAm9Ys.
\end{abstract}


\section{Introduction}

Autonomous driving technology has received more and more attention with the increasing level of automobile intelligence \cite{feng2023dense}\cite{doi:10.34133/research.0064}. However, due to the challenges of algorithm level and technology maturity, there is still a distance to the realization of high-level autonomous driving.  In fact, autonomous driving algorithms are required to undergo tens of millions of miles of data collection and testing before they can be deployed in real-world applications \cite{riedmaier2020survey}\cite{kalra2016driving}. It means that autonomous vehicles need to face enough scenarios and expose the algorithm's problems, so this process is considered fundamental for the algorithms to achieve performance improvement.

However, the actual traffic environment is complex and difficult to be exhaustive, so it is necessary to obtain diverse, typical, reasonable and interpretable environmental input information through scenario generation methods, so as to accelerate the efficiency of self-evolution and test evaluation \cite{ding2023survey}\cite{wu2024accelerated}. There have been a lot of studies on related techniques, mainly including mechanism model methods and data-driven methods.

Scenario generation based on mechanistic models aims to generate and simulate scenarios that may be encountered by autonomous vehicles by utilizing experts' understanding and professional experience of the changing rules within the scenario system, combined with logic rules and optimized solving. Rocklage et al. proposed a method for automatically generating autonomous driving scenarios for regression testing. The method defines different types of traffic scenarios by creating static and mixed scenarios to guarantee a certain coverage of parameter combinations \cite{rocklage2017automated}. Gelder et al. proposed a scenario parameter generation method and a scenario representativeness metric. This method can determine enough parameters for scenario description, and generate real parameter values by estimating the probability density function of these parameters. A Wasserstein distance based scenario representativeness metric is also proposed to quantify the realism of the generated scenario \cite{de2022scenario}.

However, although the mechanism model-based methods can make full use of existing knowledge to generate diverse and interpretable scenarios according to the test requirements, the artificial modeling method relies on a large number of expert experiences and various simplified conditions, which limits the practical application of these techniques. Compared with the mechanistic modeling methods, the data-driven methods have an obvious advantage because they can mine the potential laws from the data and fully consider the nonlinearities, uncertainties, and characteristic probability distributions in scenario generation \cite{feng2021intelligent}\cite{doi:10.34133/2020/8757403}.

Data-driven methods can be mainly divided into two categories, including natural scenario generation and adversarial scenario generation. Natural scenario generation aims to train a scenario generation model that is consistent with real driving conditions based on massive natural driving data through Bayesian network \cite{wheeler2015initial}, deep learning \cite{demetriou2023deep} and other methods. Diverse scenarios are generated by data-driven methods for training and testing of autonomous driving algorithms \cite{feng2023trafficgen}.

However, in natural scenarios, the probability of accidents for autonomous driving algorithms is very low, so it is necessary to construct targeted scenarios through adversarial scenario generation methods to approach the performance limit of algorithms quickly. Zhang et al. proposed a metamorphic testing framework for autonomous driving systems based on generative adversarial network, which is combined with real-world weather scenarios to generate driving scenarios under various extreme weather conditions \cite{zhang2018deeproad}. Jia et al. proposed a dynamic scenario generation method based on conditional generative adversarial imitation learning, in which scenario class labels are incorporated into the model to generate different types of traffic scenarios for inferring the weaknesses of the under test autonomous vehicles \cite{jia2024dynamic}. Liu proposed a reinforcement learning(RL) based safety-critical scenario generation method that uses risk and likelihood objectives to design a reward function to overcome the dimensional disaster problem and explore a wide range of safety-critical scenarios \cite{liu2023safety}. Hao et al. proposed an adversarial safety-critical scenario generation method based on natural human driving priors, which uses human driving priors and RL techniques to generate realistic safety-critical test scenarios covering both naturalness and adversarial \cite{hao2023adversarial}.

It can be seen that the adversarial scenario generation method based on data-driven method can more effectively find the vulnerabilities of the algorithm, and greatly accelerate the efficiency of algorithm training and testing. However, it is difficult for existing data-driven algorithms to control the results of generated scenarios. This difficulty has led to few studies that clearly define quantitative criteria for scenario difficulty in order to generate continuous, interpretable and targeted quantifiable scenarios with different difficulty levels through explicit quantitative indicators.

In view of the above problems, this paper proposes a data-driven quantitative representation method of scenario difficulty for autonomous driving based on environment agent policy search. To our knowledge, this is the first proposed data-driven approach for quantifying scenario difficulty representation. The environment agent is proposed, and a RL algorithm combined with mechanism knowledge is constructed to realize the policy search, so as to obtain the agent policy with adversarial behavior. To obtain information on the quantitative dimension of scenario difficulty, the model parameters of environment agents at different stages of the training process are extracted into adversarial policy groups to obtain agent policies with different adversarial intensity. In the simulation environment, the data generation of the scenario with different difficulty is carried out and the scenario database is constructed. The data-driven scenario difficulty quantitative representation model is constructed, and the feature correlation of scenario input information is extracted through the attention network, and finally the environment agent policy that can output different difficulty scenarios is obtained. The proposed method can generate reasonable scenarios with high discrimination, and can provide quantifiable difficulty representation without any expert logic rule design.

The contributions of this study are summarized as follows:

1. This paper proposes environment agent, combines mechanism knowledge and RL method to achieve efficient policy search, and obtains agent policies with adversarial behaviors.

2. This paper proposes a data generation method for varying difficulty scenarios, which combines the policy groups constructed by model parameters at different stages in the training process to provide information on the quantitative dimension of scenario difficulty.

3. This paper proposes a data-driven scenario difficulty quantitative representation model, and proves it can generate continuous and highly distinguishable scenarios with reasonable and quantifiable difficulty representations through result analysis.

This paper is organized as follows. The proposed framework is introduced in Section 2. The environment agent policy search is introduced in Section 3. The descriptions of data generation of scenarios with varying difficulty are proposed in Section 4. The details of quantitative representation of  are proposed in Section 5. In Section 6, our method is verified and compared in simulation, and section 7 concludes this paper.

\section{Proposed Framework}\label{sec:overall}

\begin{figure*}[ht!]
\centering
\begin{tabular}{c}
\includegraphics[width=0.95\textwidth]{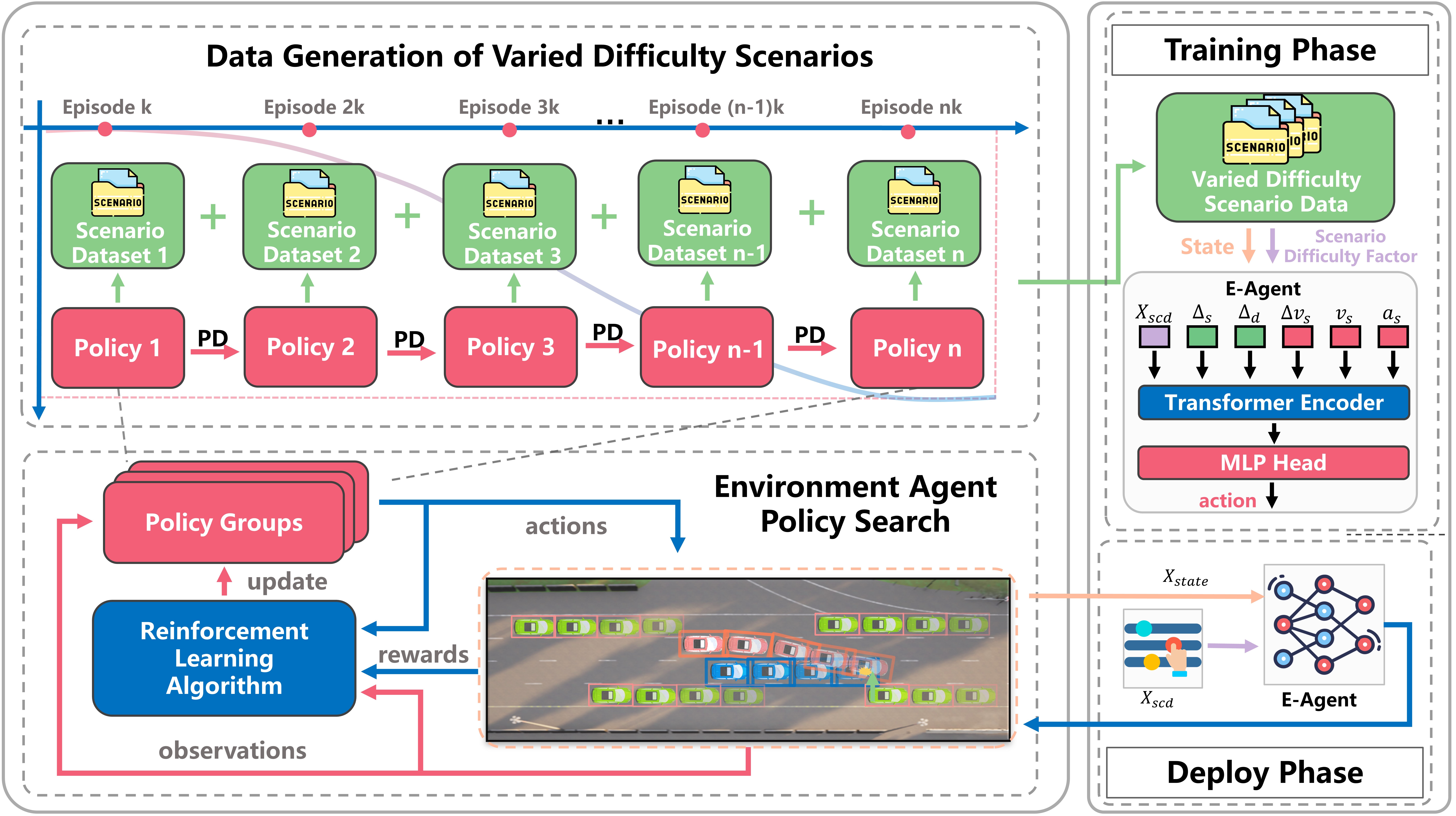}
\end{tabular}
\caption{Overall architecture of data driven quantitative representation method of scenario difficulty for autonomous driving based on environment agent policy search.}
\label{overall_architecture}
\end{figure*}

The proposed data-driven scenario difficulty quantitative representation method aims to generate reasonable scenario with high discrimination and quantifiable difficulty representation without any expert logic rule design. The main components of the proposed method are depicted in Fig. \ref{overall_architecture}, including environment agent policy search, data generation of varied difficulty scenarios and the construction of data driven quantitative representation model of scenario difficulty. The model input is the information of the surrounding traffic environment perceived by the environmental agent, as well as the continuous scenario difficulty value, and the output is the action of the environmental agent

This method employs a deep neural network as the environment agent and addresses the problem of adversarial policy search through a reinforcement learning algorithm. The model parameters trained at various stages are updated, saved, and outputted into the constructed policy group. The process of data generation varying difficulty scenarios involves two parts: deploying adversarial policies from the policy group and constructing the scenario database. Multiple policies from the policy group are successively deployed to the simulation environment for data generation of varying difficulty scenarios and appended to the scenario database. The proposed method mainly includes two phases: training phase and deployment phase. In the training phase, the attention mechanism model is used to extracts the feature association between the scenario difficulty factor and the state input. This allows for the creation of an environment agent policy capable of outputting different difficulty scenarios. And in the deployment phase, environment agent is deployed to generate scenarios with continuous difficulty levels.

\section{Environment Agent Policy Search}\label{sec:EAPS}

\subsection{Problem Definition}

Adversarial scenario generation aims to construct various complex, dangerous or extreme traffic scenarios in order to expose the weaknesses of autonomous driving algorithms when facing different environmental inputs. Based on this, improving the performance of the algorithms based on the evaluation results can further improve the reliability and safety of the autonomous driving system.

In this paper, the concept of environment agent is presented for generating adversarial scenarios. The optimization objective of the environment agent policy is to maximize the influence of the scenario information input on the performance of the ego vehicle through the agent policy search. The optimization problem can be formulated as follows:

\begin{equation}
\label{Eq_1}
{\theta ^*} = \mathop {\arg \max }\limits_\theta  \left[ {F\left( {{\pi _A},{\pi _E}\left( \theta  \right),{S_c}} \right)} \right],\
\end{equation}
where $F\left(  \cdot  \right)$ is the performance evaluation metric, ${\pi _A}$ is the ego vehicle policy, ${\pi _E}$ is the environment agent policy with parameter $\theta$ and ${S_c}$ is the scenario dynamics. ${S_c}$ is used to characterize the state transitions of all agents.

Reinforcement learning algorithm is dedicated to searching for optimal policies in order to maximize future returns. Therefore, this paper constructs a reinforcement learning problem to realize the policy search for the above optimization problem. Considering that the adversarial behaviors of the environment agent must satisfy the basic logic rules, this paper incorporates the mechanism knowledge in order to speed up the efficiency of the policy search.

\subsection{Environment Agent Design}
\subsubsection{Markov Decision Process (MDP)}

The dynamic interaction process between the environment agent and the ego vehicle is constructed as a Markov Decision Process, which can be defined by the following five tuples:

\begin{equation}
M = (S,A,P,\pi ,R),\
\end{equation}
where $S$ is state space, $A$ is action space, $P$ is probabilistic model of state transition, $\pi$ is policy model and $R$ is reward.

The core content of MDP is the Markov property. It states that the future state of a system depends solely on its current state, independent of the sequence of states that led to the current state. In other words, the next state in the sequence is completely determined by the current state and is not affected by the sequence of previous states. The Markov property can be expressed as:

\begin{equation}
\begin{array}{*{20}{c}}
{p\left( {{s_{t + 1}}\mid {s_t}} \right) = p\left( {{s_{t + 1}}\mid {h_t}} \right)}\\
{p\left( {{s_{t + 1}}\mid {s_t},{a_t}} \right) = p\left( {{s_{t + 1}}\mid {h_t},{a_t}} \right)},
\end{array}\
\end{equation}
where the history of states is ${h_t} = \left\{ {{s_1},{s_2},{s_3}, \ldots ,{s_t}} \right\}$.

\subsubsection{Soft actor critic}

\begin{figure}[h]
\centering
\includegraphics[width=4.0in]{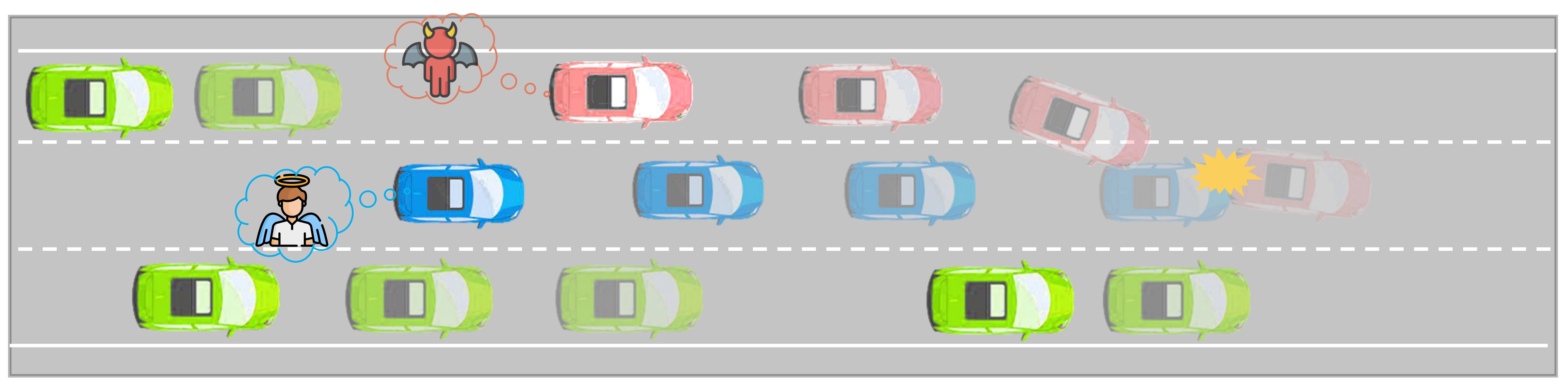}
\caption{Adversarial behavior of environment agent. The blue car represents the ego vehicle driving normally, and the red car represents the environment agent with adversarial behavior.}
\label{Adversarial_scenario}
\end{figure}

Reinforcement Learning is a powerful tool to solve MDPs and find optimal or suboptimal policies for learning agents. Soft actor critic (SAC) algorithm \cite{haarnoja2018soft} is a widely used reinforcement learning algorithm. It is a model-free, off-policy algorithm based on actor-critic framework. SAC uses stochastic policy, which can make the policy as random as possible. The agent can explore the state space S more fully, avoid the policy falling into the local optimum early, and can explore multiple feasible solutions to complete the specified task to improve the anti-interference ability.It should be noted that the proposed algorithm is not limited to the use of the SAC algorithm, but other novel reinforcement learning methods can be also applied to the method.

According to MDP, SAC seeks to solve the following maximum entropy problem:

\begin{equation}
{\pi ^ * } = \arg \mathop {\max }\limits_\pi  {_{\left( {{s_t},{a_t}} \right) \sim {\rho _\pi }}}\left[ {\mathop \sum \limits_t r\left( {{s_t},{a_t}} \right) + \alpha {\cal H}\left( {\pi \left( { \cdot \mid {s_t}} \right)} \right)} \right],\
\label{pi_star}
\end{equation}
where ${\cal H}$ is entropy, $\alpha$ is the temperature parameter. With $\alpha  \to 0$, the maximum entropy RL gradually approaches the conventional RL

The basic flow of soft policy iteration is to run two steps of policy evaluation and policy improvement alternately until convergence. In the policy evaluation step of soft policy iteration, the value of policy $\pi$ is calculated according to the maximum entropy objective. Soft Q-value can be obtained by iterating Behrman backup operator, that is:

\begin{equation}
{{\cal T}^\pi }Q\left( {{s_t},{a_t}} \right) \buildrel \Delta \over = r\left( {{s_t},{a_t}} \right) + \gamma {_{{{\bf{s}}_{t + 1}} \sim p}}\left[ {V\left( {{s_{t + 1}}} \right)} \right]\
\end{equation}

The goal of soft policy improvement is to search for a new policy ${\pi _{new}}$ that is better than the current policy ${\pi _{old}}$. To achieve this goal, we can represent the policy as a gaussian distribution and reduce the gap between the current policy and the new policy by minimizing the Kullback-Leibler divergence.

\begin{equation}
{\pi _{{\rm{new}}}} = \arg \mathop {\min }\limits_{\pi ' \in {\rm{\Pi }}} {D_{{\rm{KL}}}}\left( {\pi '\left( { \cdot \mid {{\bf{s}}_t}} \right)\parallel \frac{{\exp \left( {{Q^{{\pi _{{\rm{old}}}}}}\left( {{{\bf{s}}_t}, \cdot } \right)} \right)}}{{{Z^{{\pi _{{\rm{old}}}}}}\left( {{{\bf{s}}_t}} \right)}}} \right),\
\end{equation}
where ${{Z^{{\pi _{{\rm{old}}}}}}\left( {{{\bf{s}}_t}} \right)}$ is the normalized distribution of $Q$ values, it will not contribute to the policy gradient, so it can be ignored.

Soft policy iteration has convergence and optimality, see \cite{haarnoja2018softt}.

\subsubsection{State and action space design}

During the normal operation of autonomous vehicles, sudden cut-in of environment vehicles is a common risky behavior that may lead to emergency situations and jeopardize traffic safety. First, sudden cut-in may disrupt the autonomous vehicle's path and speed planning, requiring the system to react quickly to avoid collisions or violations. Second, sudden engagement can trigger emergency braking or evasive maneuvers, and the braking system and evasive strategies need time to respond, potentially resulting in collisions with rear-end vehicles or roadway blockages that can lead to traffic congestion and more serious accidents.

Fig. \ref{Adversarial_scenario} presents a schematic of the environment agent generating adversarial behavior. Among them, the blue car represents the ego vehicle and the red car represents the environment agent. When the ego vehicle is driving normally, the environment agent will look for a time to cut in from the neighboring lanes to the current lane, and minimize the ego vehicle’s performance as much as possible through the control of throttle, brake, and steering, so as to maximize the risk of side collision, corner collision or front collision.

\begin{figure}[h]
\centering
\includegraphics[width=4.0in]{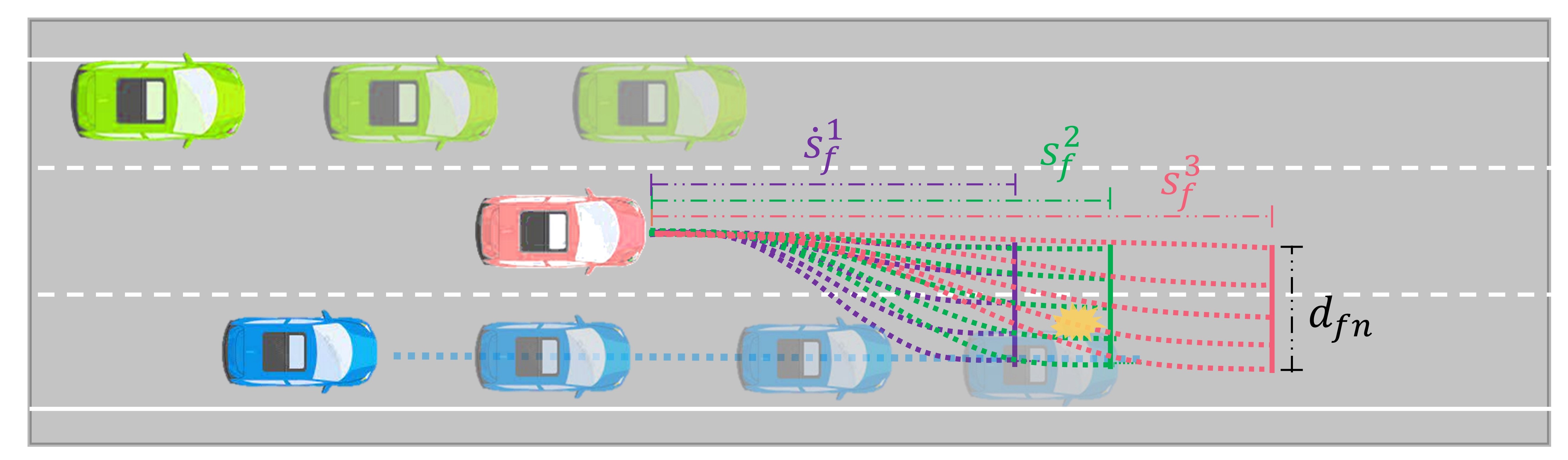}
\caption{Construction of adversarial policy search problem for environment agent based on polynomial curves.}
\label{quintic_polynomial}
\end{figure}

The state design is required to fully consider the input information required by the environment agent. Define the state space $S$ as follows:

\begin{equation}
S = \left[ {\begin{array}{*{20}{c}}
{\Delta s}&{\Delta d}&{\Delta {v_s}}&{{v_s}}&{{a_s}}
\end{array}} \right],\
\end{equation}
where $\Delta s$ is the relative longitudinal distance between the environment agent and the ego vehicle, $\Delta d$ is the relative lateral distance, $\Delta {v_s}$ the relative longitudinal velocity, ${v_s}$ and $a$ are the longitudinal velocity and the longitudinal acceleration of the environment agent, respectively.

The underlying principle of the adversarial policy generated by the environment agent is that the agent should be as close to the ego vehicle as possible to maximize the risk of collision. Therefore, the behavior toward the environment agent moving away from the ego vehicle is against the basic rules of logic. To solve this problem, this paper combines the trajectory planning method based on mechanism rules to construct the reinforcement learning problem and the action space.

As shown in Fig. \ref{quintic_polynomial}, the quintic and quartic polynomial curves are respectively applied to describe the longitudinal and lateral trajectory planning process of environment agent \cite{wu2018time}:

\begin{equation}
\label{cal_polynomials 1}
\left[ {\begin{array}{*{20}{c}}
1&0&0&0&0&0\\
0&1&0&0&0&0\\
0&0&2&0&0&0\\
1&t&{{t^2}}&{{t^3}}&{{t^4}}&{{t^5}}\\
0&1&{2t}&{3{t^2}}&{4{t^3}}&{5{t^4}}\\
0&0&2&{6t}&{12{t^2}}&{20{t^3}}
\end{array}} \right] \cdot {p_d} = \left[ {\begin{array}{*{20}{c}}
{{d_0}}\\
{{{\dot d}_0}}\\
{{{\ddot d}_0}}\\
{{d_{fn}}}\\
{{{\dot d}_{fn}}}\\
{{{\ddot d}_{fn}}}
\end{array}} \right],\
\end{equation}

\begin{equation}
\label{cal_polynomials 2}
\left[ {\begin{array}{*{20}{c}}
1&0&0&0&0\\
0&1&0&0&0\\
0&0&2&0&0\\
0&1&{2t}&{3{t^2}}&{4{t^3}}\\
0&0&2&{6t}&{12{t^2}}
\end{array}} \right] \cdot {p_s} = \left[ {\begin{array}{*{20}{c}}
{{s_0}}\\
{{{\dot s}_0}}\\
{{{\ddot s}_0}}\\
{{{\dot s}_f}}\\
{{{\ddot s}_f}}
\end{array}} \right],\
\end{equation}

\begin{equation}
{p_d} = {\left[ {\begin{array}{*{20}{c}}
{a_d^0}&{a_d^1}&{a_d^2}&{a_d^3}&{a_d^4}&{a_d^5}
\end{array}} \right]^T},\
\end{equation}

\begin{equation}
{p_s} = {\left[ {\begin{array}{*{20}{c}}
{a_s^0}&{a_s^1}&{a_s^2}&{a_s^3}&{a_s^4}
\end{array}} \right]^T},\
\end{equation}
where $t$ is time variable, $p_d$ is coefficients of quintic polynomials for lateral planning, $[ {\begin{array}{*{20}{c}}
{{d_0}}&{{{\dot d}_0}}&{{{\ddot d}_0}}&{{d_{fn}}}&{{{\dot d}_{fn}}}&{{{\ddot d}_{fn}}}
\end{array}} ]^T$ is the boundary condition of a quintic polynomial, $p_s$ is coefficients of quartic polynomials for longitudinal planning and $[{\begin{array}{*{20}{c}}
{{s_0}}&{{{\dot s}_0}}&{{{\ddot s}_0}}&{{{\dot s}_f}}&{{{\ddot s}_f}}
\end{array}}]^T$ is the boundary condition of a quartic polynomial. In the Frenet coordinate system \cite{werling2010optimal}, $s$ represents the arc length along the road, used for longitudinal position description, while the lateral position is represented by the offset \( d \) perpendicular to the path. The polynomial coefficients $p_d$ and $p_s$ can be solved by substituting the planning time \( T_c \) into the polynomial planning equations. Specifically, the parameters \( d_0 \), \( \dot{d}_0 \), \( \ddot{d}_0 \) represent the initial position, initial velocity, and initial acceleration, respectively; \( d_{fn} \), \( \dot{d}_{fn} \), \( \ddot{d}_{fn} \) represent the final position, final velocity, and final acceleration, respectively. Similarly, the parameters \( s_0 \), \( \dot{s}_0 \), \( \ddot{s}_0 \) represent the initial position, initial velocity, and initial acceleration, respectively; \( s_f \), \( \dot{s}_f \), \( \ddot{s}_f \) represent the final position, final velocity, and final acceleration, respectively. \( a_d^i \) (i=0, 1, 2, 3, 4, 5) are the coefficients of the quintic polynomial, and \( a_s^i \) (i=0, 1, 2, 3, 4) are the coefficients of the quartic polynomial. Using the initial and final boundary conditions in the above matrix equations, the polynomial coefficients can be solved to obtain the polynomial forms for trajectory planning. The quintic polynomial is used for lateral planning, and the quartic polynomial is used for longitudinal planning.

The design of action space  needs to consider the goal of the autonomous driving task. The action space of the reinforcement learning problem is designed as follows:

\begin{equation}
\textnormal{A} = \left[ {\begin{array}{*{20}{c}}
{{d_{fn}}}&{{{\dot s}_f}}
\end{array}} \right],\
\end{equation}
where ${d_{fn}}$ and ${\dot s_f}$ are the expected lateral offset and expected longitudinal velocity after time ${T_c}$. A simple PID controller is applied for tracking the desired trajectory generated by the reinforcement learning algorithm \cite{farag2020complex}.

\subsubsection{Reward design}

The design of reward function is very important for reinforcement learning algorithms. The reward function is used to guide and evaluate the agent’s behavior, which has a direct impact on the performance of the algorithm. In this paper, the reward function with adversarial nature is designed to approach the performance boundary of the ego vehicle by guiding the environment agent to continuously generate adversarial behaviors.

Firstly, the adversarial policy of the environment agent should improve the collision risk with the ego vehicle as much as possible. Artificial potential field method is a kind of virtual force method \cite{khatib1986real}, the basic idea of which is to design the movement of the robot in the surrounding environment into an abstract artificial gravitational field. The artificial potential Field contains two kinds of force fields: the attractive field formed by the position of the moving target and the repulsive field formed by the obstacle. Therefore, inspired by the improved artificial potential field theory \cite{shi2009global}, the collision risk of the ego vehicle is quantified.

The risk reward $r_1$ is designed as follows:

\begin{equation}
\begin{aligned}
{r_1} =  - \sum\limits_{{\rm{i}} = 1}^m {\min \left( {\exp \left( {\frac{{(\min {{(\max (\Delta d - {L_a},{r_{\min }}),{r_{\max }})}^2}}}{{\delta _1^2}} + \frac{{(\min {{(\max (\Delta s - {L_b},{r_{\min }}),{r_{\max }})}^2}}}{{\delta _2^2}}} \right) \cdot {k_f} \cdot g + b,0} \right)} \
\end{aligned}
\end{equation}
where $r_{\min }$ and $r_{\max }$ are the safe distance penalty lower bound constant and the safe distance penalty upper bound constant, respectively. $\delta _1$ and $\delta _2$ are the lateral repulsion influence factor and the longitudinal repulsion influence factor, respectively. $L_a$ is the vehicle width, $L_b$ is the vehicle length, $k_f$ is the scale factor. In order to keep the reward value in a reasonable range, $g$ is defined as the linear mapping proportion and $b$ is defined as the linear mapping deviation. $g$ and $b$ can be expressed as follows:

\begin{equation}
g =  - \frac{{{\rho _n}}}{{{k_f} \cdot S - {k_f}}},\
\end{equation}

\begin{equation}
b = {\rho _n} - g \cdot {k_f},\
\end{equation}
where ${\rho _n}$ is the safe distance penalty factor. $S$ is the proportional correction factor, which can be expressed as:

\begin{equation}
S = \exp (\frac{{{{({d_{thre}} - {L_a})}^2}}}{{\delta _1^2}} + \frac{{{{({s_{thre}} - {L_b})}^2}}}{{\delta _2^2}}),\
\end{equation}
where ${d_{thre}}$ and ${s_{thre}}$ are the lateral and longitudinal threshold value of safe distance penalty, respectively.

The vehicle speed is encouraged to be maintained within a reasonable range. In the start-up stage, the reward function is designed to guide the vehicle to accelerate from $0$, while in the speed maintenance stage, the reward function is designed to maintain in a preset speed interval. In order to design the adversarial reward to harm the longitudinal performance of the vehicle, the minimum speed reward $r_2$ is designed as follows:

\begin{equation}
{r_2} =  - {\rho _L} \times {\kappa_L} - {\rho _H} \times {\kappa_H},\
\end{equation}
where ${\rho _L}$ is the start-up stage reward factor, ${\rho _H}$ is the speed maintenance stage reward factor, ${\kappa_L} = \frac{{{v_s}}}{{{v_{\min }}}}$ is the ratio between the longitudinal velocity $v_s$ and the desired minimum velocity $v_{min}$, ${\kappa_H} = \frac{{{v_s} - {v_{\min }}}}{{{v_{\max }} - {v_{\min }}}}$ is the ratio of the vehicle speed $v_s$ to the desired maximum speed $v_{max}$ based on the desired minimum speed $v_{min}$.

\begin{table}[!t]
    \caption{Reward function parameters setting\label{tab:Reward function parameters setting}}
    \centering
    \begin{tabular}{cc}
            \hline
            Reward function parameters terms & Symbol \& Value \\
            \hline
            Safe distance penalty lower bound constant  & ${r_{\min }} = 0.0$ \\
            Safe distance penalty upper bound constant  & ${r_{\max }} = 150.0$ \\
            Lateral repulsion influence factor & ${\delta _1} = 8.0$ \\
            Longitudinal repulsion influence factor & ${\delta _2} = 10.0$ \\
            Scale factor  & ${k_f} = 0.001$ \\
            Vehicle width & ${L_a} = 2.077m$ \\
            Vehicle length & ${L_b} = 5.037m$ \\
            Safe distance penalty factor & ${\rho _n} =  - 18.0$ \\
            Lateral threshold value of safe distance penalty & ${d_{thre}} = 0.8$ \\
            Longitudinal threshold value of safe distance penalty & ${S_{thre}} = 20.0$  \\
            Start-up stage reward factor & ${\rho _L} = 0.5$ \\
            Speed maintenance stage reward factor & ${\rho _H} = 4.0$ \\
            Expected minimum speed & ${v_{\min }} = 7.5m/s$ \\
            Expected maximum speed& ${v_{\max }} = 22.0m/s$ \\
            Collision reward factor& ${\rho _{coll}} = 200.0$ \\
            \hline
            \end{tabular}
\end{table}

Causing the collision accident is considered to be the most serious consequence that can result from the adversarial behavior produced by the environment agent. Therefore, when the collision occurs, the collision reward $r_3$ is set as follows.

\begin{equation}
{r_3} = \left\{ {\begin{array}{*{20}{c}}
0&{\begin{array}{*{20}{c}}
{not}&{collision}
\end{array}}\\
{{\rho _{coll}}}&{collision}
\end{array}}, \right.\
\end{equation}
where ${\rho _{coll}}$ is collision reward factor. ${\rho _{coll}}$ is set to a positive value to indicate that policies that receive this reward are encouraged.

The total reward function is:

\begin{equation}
R = -{r_1} + {r_2} + {r_3}\
\end{equation}

All reward function parameters are listed in Table I.

\section{Data Generation of Scenarios with Varying Difficulty}

In Section \ref{sec:EAPS}, we introduce the concept of environment agent to realize the adversarial policy search by combining logic rules with reinforcement learning. However, due to the black-box nature of data-driven methods, while adversarial actions can be generated, the difficulty of generating adversarial actions is difficult to quantify accurately, which limits the rationality of adversarial scenario generation.

In this section, a data generation method based on scenarios with varying difficulty is presented. The method uses the performance of different stages in the policy search convergence process as a reference to quantify the adversarial intensity, thereby achieving a quantitative representation of scenario difficulty. The model parameters of the environment  agent trained on different stages are updated and saved, and then output to the constructed policy group. The policy group is used to generate data that forms the basis for training the scenario difficulty quantitative representation model.

\subsection{Policy Group Construction}

\begin{figure}[h]
\centering
\includegraphics[width=4.5in]{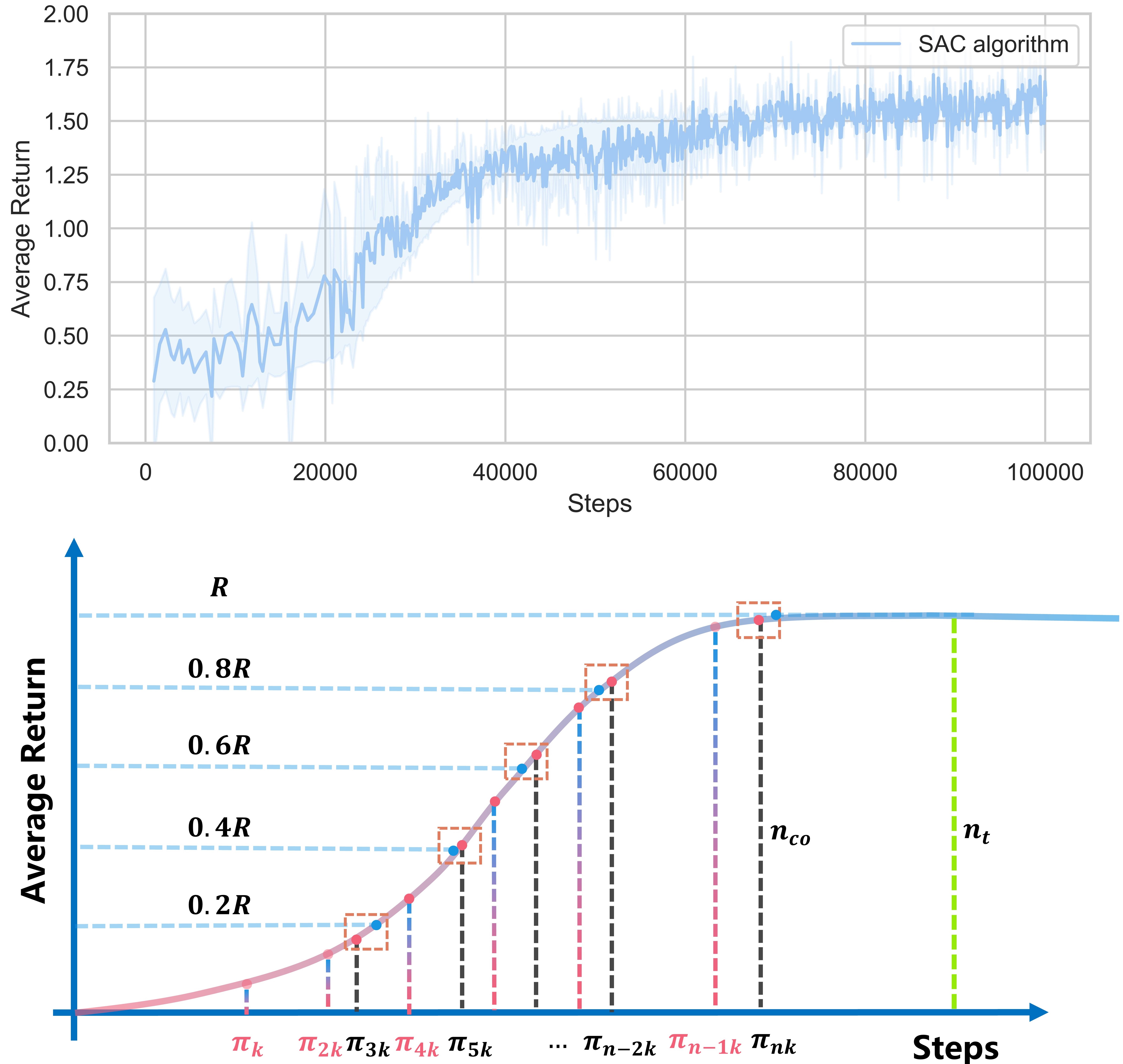}
\caption{Construction of policy group.}
\label{policy_group}
\end{figure}

As shown in Fig. \ref{policy_group}, the schematic of the average return during the policy search of the environment agent is presented. The horizontal axis shows the number of training steps and the vertical axis shows the average return. The training details can be seen in Section \ref{sec:Results}.

A reinforcement learning training process with stable convergence can be divided into two phases, i.e., the performance improvement phase and the convergence stabilization phase. In the performance improvement phase, the average return is still continuously increasing, which indicates that the policy search is still ongoing and the model parameters are still being updated to peruse better performance. In the convergence stabilization phase, however, the average return remains basically unchanged, indicating that the policy search is basically over, and the obtained policy is already the optimal policy that the current algorithm can achieve.

Considering that the intensity of the environment agent's adversarial behavior against the ego vehicle may increase nonlinearly with the training process, a policy filtering method is proposed  to ensure that the policies within the policy group can generate distinguishable and reasonable adversarial scenarios with different intensities. The policy filtering algorithm is shown in Algorithm \ref{alg:alg1} and Fig. \ref{policy_group}.

The initialization and parameter setting of the algorithm are performed in Line 1-Line 3. In Line 4-Line 5, through trial training of the environment agent, reasonable total steps ${n_t}$, performance improvement steps ${n_{co}}$, and maximum average return $R$ are obtained.

The policy search of the environment agent ${\pi _E}$ is shown In Line 7, and the parameter $\theta$ of ${\pi _E}$ is updated using Eq. \ref{pi_star}. Line 8 - Line 12 represent the sampling operations conducted on model parameters during the convergence process of the algorithm. The sampled model parameters are placed in the set $\Xi$. The sampling interval is $k$, i.e., the model is sampled every $k$ rounds. The sampled model ${\pi _k}$ and the corresponding training steps ${e_k}$ form a binary group and are placed in the set $\Xi$, which can be denoted as:

\begin{equation}
\Xi  = \left\{ {{\Pi _k},{\Pi _{2k}},{\Pi _{3k}},...,{\Pi _{nk}}} \right\},\
\end{equation}
where ${\Pi _k} = \{ {\pi _k},{e_k}\}$, $n = len\left( \Xi  \right)$.

As shown in Fig. \ref{policy_group}, the bottom of the horizontal axis represents the sampling models ${\pi _k},{\pi _{2k}},...,{\pi _{nk}}$ at different phases in the set $\Xi $, corresponding to the positions at different time steps ${e_k},{e_{2k}},...,{e_{nk}}$.

\begin{algorithm}[H]
\caption{Policy Group Construction}\label{alg:alg1}
\begin{algorithmic}[1]
\STATE \textbf{Input:} Ego vehicle policy ${\pi _A}$, scenario dynamics ${S_c}$;
\STATE \textbf{Output:} Total steps ${n_t}$, performance enhancement steps ${n_{co}}$, maximum average return $R$
\STATE  Initialize Policy group $O = \emptyset$, set $\Xi  = \emptyset$, environment agent network ${\pi _E}$ with weights $\theta$, episode number $\vartheta  = 0$;
\STATE Set parameters Sampling interval $k$;
\STATE  Execute Trial train environment agent ${\pi _E}$ under ${\pi _A}$ and ${S_c}$
\STATE \textbf{for} \emph{i=1, 2, ...,}${n_{t}}$ \textbf{do}
\STATE \hspace{0.5cm}\textbf{Execute} policy search, update environment agent network ${\pi _E}$ using Eq. \ref{pi_star}
\STATE \hspace{0.5cm}\textbf{if} $i < {n_{co}}$ \textbf{and} $\bmod \left( {\vartheta ,k} \right) = 0$ \textbf{then}
\STATE \hspace{1.0cm}${e_k = i}$
\STATE \hspace{1.0cm}Collect ${\Pi _k} = \left( {{\pi _k},{e_k}} \right)$
\STATE \hspace{1.0cm}$\Xi  \leftarrow \Xi  \cup \left\{ {{\Pi _k} = \left( {{\pi _k},{e_k}} \right)} \right\}$
\STATE \hspace{0.5cm}\textbf{if} episode end \textbf{then}
\STATE \hspace{1.0cm}$\vartheta  = \vartheta  + 1$
\STATE \textbf{for} \emph{i=1, 2, 3, 4, 5} \textbf{do}
\STATE \hspace{1.0cm}$n_l^i = {\Gamma ^{ - 1}}\left( {0.2i \cdot R} \right)$
\STATE \hspace{0.5cm}\textbf{for} \emph{j=1, 2, ...,}$len(\Xi )$ \textbf{do}
\STATE \hspace{1.5cm}$\tau  = \arg \min \left( {n_l^i - \Xi [j][2]} \right)$
\STATE \hspace{1.5cm}$O \leftarrow O \cup \Xi [\tau ]$
\end{algorithmic}
\label{alg1}
\end{algorithm}

In Line 13-Line 17, policy filtering is performed to ensure that the intensity of the adversarial behaviors produced by the environment agent models in policy group $O$ grows linearly. To quantify the intensity of adversarial strength, maximum average return $R$ is used. Fig. \ref{policy_group} illustrates the policy filtering process. First, the mapping relation between the average return curve and the training steps is defined as:

\begin{equation}
{R_s} = \Gamma \left( {{n_s}}\right), \
\end{equation}
where ${R_s}$ is the corresponding average return value at $n_s$ steps.

In Line 13-Line 14, the maximum average return $R$ is linearly split, as shown in Fig. \ref{policy_group}, to obtain the number of training steps corresponding to the dividing line (the horizontal coordinates corresponding to the blue circle center point). In Line 15-Line 16, a find algorithm is executed to find the index $\tau$ of the sampled model that is closest to $n_l^i$ in ${e_k},{e_{2k}},...,{e_{nk}}$ (i.e. closest pink center point to the blue center point in Fig. \ref{policy_group}).

In Line 17, the sampling models are extracted from the set $\Xi$ according to the index $\tau$ and deposited into $O$ to finalize the construction of the policy group. The sampled models that have been filtered and deposited into the policy group $O$ are shown as the black dashed lines in Fig. \ref{policy_group}.

\subsection{Varied Difficulty Scenario Dataset}

The constructed policy group $O$ is used to build a varied difficulty scenario dataset and this process is carried out in a simulation environment. The parameters of the simulation environment are randomly configured, which is to simulate more diverse real-world inputs and ensure the generalization of subsequent model training.

The pseudocode of the construction for the varied difficulty scenario dataset is shown inm{Algorithm \ref{alg:alg2}.

\begin{algorithm}[H]
\caption{Varied Difficulty Scenario Dataset Construction}\label{alg:alg2}
\begin{algorithmic}[1]
\STATE \textbf{Input:} Policy group $O$, traffic model ${\tau _{tf}}$ with parameters $\sigma$;
\STATE  Initialize simulation environment ${S_{world}}$, varied difficulty scenario dataset $\Psi$;
\STATE  Set parameters same category data size ${n_{sc}}$, episode number ${n_{episode}}$;
\STATE \textbf{for} \emph{i=1, 2, ...,}$len(O)$ \textbf{do}
\STATE \hspace{0.5cm}\textbf{for} \emph{j=1, 2, ...,}${n_{episode}}$ \textbf{do}
\STATE \hspace{1.0cm}\textbf{Deploy} the policy $O\left[i \right]$ to the environment agent
\STATE \hspace{1.0cm}\textbf{Random choose} traffic model parameters $\sigma$
\STATE \hspace{1.0cm}\textbf{Generate} random traffic flow using ${\tau _{tf}}$ in simulation
\STATE \hspace{1.0cm}Using environment agent to collect $\left( {S,0.2*i,\textnormal{a}} \right)$
\STATE \hspace{1.0cm}$\Psi  \leftarrow \Psi  \cup \left( {S,0.2*i,\textnormal{a}} \right)$
\STATE \hspace{1.0cm}\textbf{if} steps num $ > i*{n_{sc}}$ \textbf{then}
\STATE \hspace{1.5cm}break
\end{algorithmic}
\label{alg2}
\end{algorithm}

First, Line 1 initializes the simulation environment ${S_{world}}$ and varied difficulty scenario dataset $\Psi$. Line 2 - Line 3 determine the input and parameter setting of the algorithm. line 4 - Line 15 represent a complete dataset construction process. In Line 6, the policies in the policy group $O$ are deployed to the environment agent. In Line 7 - Line 8, the random traffic flow ${\tau _{tf}}$ is generated in the simulation environment by randomizing the parameters of the traffic model $\sigma$. In Line 9 - Line 10, the environment agent is used to collect the ternary $\left( {S,0.2*i,\textnormal{a}} \right)$ into $\Psi$, where $0.2*i$ denotes the difficulty factor of the scenario. Lines 11 to 12 indicate that when the collected data size of the same category exceeds ${n_{sc}}$, the data collection for the corresponding difficulty level finished. The algorithm then jump out the current inner loop, proceeds to deploy the environment agent's policy for the next difficulty level, and continues until varied difficulty scenario datasets $\Psi$ are completely collected.

\section{Quantitative Representation of Scenario Difficulty}

To obtain the environment agent model with adversarial behavior, the quantitative representation model is constructed. The model is based on deep neural network with transformer decoding architecture. As shown in Fig. \ref{overall_architecture}, the input of the model is the state $s$ and the scenario difficulty factor ${X_{scd}}$, and the output is the action $a$ of the environment agent.

In recent years, the transformer model has gained much attention \cite{hu2023planning}\cite{vaswani2017attention}, and its core idea is the self-attention mechanism. The key to the self-attention mechanism is to assign different weights to certain positions, thus enabling the model to better capture long-distance dependencies and global information in the input sequence. This mechanism not only contributes to a deeper understanding of the intrinsic structure of the sequence, but also provides interpretability for the model, allowing it to focus on important parts of the sequence.

The proposed quantitative mode is constructed based on the transformer encoder model \cite{dosovitskiy2020image}. The transformer encoder consists of alternating layers of  self-attention and Multi-Layer Perceptron (MLP) block. Layer norm (LN) is applied before every block, and residual connections after every block.

The input of the neural network includes the scenario difficulty factor ${X_{scd}}$ and the relative information between the environment agent and the ego vehicle. The input vector can be expressed as follows.

\begin{equation}
{\rm{X}} = \left[ {{X_{scd}},s} \right] = \left[ {{X_{scd}},{\Delta _s},{\Delta _d},\Delta {v_s},{v_s},{a_s}} \right]\
\end{equation}

The standard Transformer model takes a one-dimensional sequence of token embeddings as input. The embedding projection is then applied to the extracted features from the input vector.

\begin{equation}
{h_0} = \left[ {{X_{scd}}E;{\Delta _s}E;{\Delta _d}E,\Delta {v_s}E,{v_s}E,{a_s}E} \right],\begin{array}{*{20}{c}}
{}&{E \in {^{P \times D}}}
\end{array}\
\end{equation}

The transformer encoder model can be expressed as:

\begin{equation}
{\bf{z}}_\ell ^\prime  = {\mathop{\rm MSA}\nolimits} \left( {{\mathop{\rm LN}\nolimits} \left( {{{\bf{z}}_{\ell  - 1}}} \right)} \right) + {{\bf{z}}_{\ell  - 1}}\begin{array}{*{20}{c}}
{}&{\ell  = 1...L}
\end{array}\
\label{MSA}
\end{equation}

\begin{equation}
{\bf{y}} = {\rm{LN}}\left( {{\bf{z}}_L^0} \right)\
\end{equation}

In Eq. \ref{MSA}, the self-attention criterion divides the input embedding into three vectors $V$, $K$ and $Q$ \cite{li2022lane}. The scaled dot-product attention is calculated according to Eq. \ref{QKV}.

\begin{equation}
\Theta  = Attention(Q,K,V) = {\mathop{\rm softmax}\nolimits} \left( {\frac{{Q{K^T}}}{{\sqrt {{d_k}} }}} \right)V,\
\label{QKV}
\end{equation}
where $\Theta$ is scores matrix, $Q$ is a query vector, $K$ is a key vector, $V$ is a value vector, and ${d_k}$ is a normalization.

The training process of scenario difficulty quantitative representation model $\varphi$ is shown in Algorithm \ref{alg3}. $\varphi$ is trained with data from varied difficulty scenario dataset $\Psi$. In line 13, a batch $B$ is randomly sampled from $\varphi$. In line 14 and line 15, the loss function is calculated by the mean square error (MSE) between the predicted and true values, where $a$ and $\hat a$ denote the ground truth and predicted values, respectively. The model is trained using the Adams optimizer.

The scores matrix of transformer can be used to analyze the focus of the model when processing sequences, thus improving the interpretability of the model's behavior. The scores matrix $\Theta$ can calculated using Eq. \ref{QKV}.

\begin{algorithm}[H]
\caption{Training process of  scenario difficulty quantitative representation model}\label{alg:alg3}
\begin{algorithmic}[1]
\STATE \textbf{Input:} Varied difficulty scenario dataset $\Psi$;
\STATE Initialize quantitative representation model $\varphi$ with weights ${\theta _{sd}}$;
\STATE Set parameters batch size ${n_b}$, epoch number ${n_{epoch}}$;
\STATE \textbf{for} \emph{i=1, 2, ...,}${n_{epoch}}$ \textbf{do}
\STATE \hspace{0.5cm}\textbf{Random sample} $B = {\left( {{X_{scd}},s} \right)_{i = 0:{n_b} - 1}}$ from $\Psi$
\STATE \hspace{0.5cm}Predict $\hat a = {\varphi _{{\theta _{sd}}}}\left( {{X_{scd}},s} \right)$
\STATE \hspace{0.5cm}Compute MSE loss ${\cal L} = {\left\| {a - \hat a} \right\|^2}$
\STATE \hspace{0.5cm}Update ${\theta _{sd}}$ by ${\hat \nabla _{{\theta _{sd}}}}{{\cal L}_\varphi }({\theta _{sd}})$ using Adams optimizer
\STATE \textbf{end for}
\end{algorithmic}
\label{alg3}
\end{algorithm}

\begin{equation}
\Theta  = \left[ {\begin{array}{*{20}{c}}
{\mu _{scd}^{1 \times 1}}&{\mu _{{S_{[1]}}}^{1 \times 2}}& \cdots &{\mu _{{S_{[n]}}}^{1 \times (n + 1)}}\\
{\mu _{{S_{[1]}}}^{2 \times 1}}&{\mu _a^{2 \times 2}}& \cdots &{\mu _a^{2 \times (n + 1)}}\\
 \vdots & \vdots & \ddots & \vdots \\
{\mu _{{S_{[n]}}}^{n + 1 \times 1}}&{\mu _a^{n + 1 \times 2}}& \cdots &{\mu _a^{n + 1 \times (n + 1)}}
\end{array}} \right],\
\end{equation}
where $\mu$ is a scalar, and the bottom-right symbol is used to distinguish the source of $\mu$. $\Theta$ can be used to characterize the feature correlation between each state. For the scenario difficulty quantitative representation model, the state correlation between the scenario difficulty factor ${X_{scd}}$ and the relative information between the environment agent and the ego vehicle is the most noteworthy. By summing the elements along columns in the matrix $\Theta$, a new matrix ${\tilde \Theta ^{1 \times (n + 1)}}$ can be obtained.

In order to obtain an ordering of the characteristic correlation between the scenario difficulty factor ${X_{scd}}$ and the other states $s$, ${\tilde \Theta ^{1 \times (n + 1)}}$ is sorted and the indexes of the array elements sorted are obtained, as shown in Eq. \ref{sort}:

\begin{equation}
\Lambda  = argsort({\tilde \Theta ^{(n + 1) \times (n + 1)}}),\
\label{sort}
\end{equation}
where the index matrix $\Lambda$ is the vector of $1 \times n$ and the corresponding sorted feature contribution value is $\Upsilon  = {\tilde \Theta ^{1 \times (n + 1)}}\left[ \Lambda  \right]$.

\section{Results}\label{sec:Results}

\subsection{Experiment Setting}

In this section, the proposed data driven quantitative representation method of scenario difficulty for autonomous driving is validated in the simulation platform. The training scenario consists of randomly generated traffic flows with speeds ranging from 8 m/s to 12 m/s within 180 m of the ego vehicle, as well as the surrounding vehicle with adversarial behavior defined as the environmental agent. The training environment is constructed in the simulation software CARLA \cite{dosovitskiy2017carla}.

The soft actor algorithm consists of the value network and the policy network. The value network is a fully connected neural network with three layers, including 5 input neurons, 1 output neuron, and 256 hidden layer neurons. The policy network is a four-layer fully connected neural network with 5 input neurons, 2 output neurons, and 256 hidden layer neurons. The scenario difficulty quantitative representation model $\varphi$ consists of a transform decoder and a fully connected layer of seven layers, where the number of neurons in the fully connected layer is 512.

The parameters of the algorithm are set as shown in Table \ref{tab: Hyperparameters for the simulation}. The parameter settings are subject to detailed selection and tuning. The simulation step size $\Delta t = 0.1s$ balances computational efficiency and simulation accuracy; the value network discount factor $\gamma {\rm{  =  0}}{\rm{.99}}$ ensures that future rewards are not overly discounted, promoting long-term decision-making; the learning rate of the adjust temperature $\lambda  = 3e - 4$ and the learning rate of the target network $\tau_{1}  = 5e - 3$ are fine-tuned through preliminary experiments to ensure stability and efficient convergence; the total number of training steps ${n_t} = 100000$ is determined based on the task complexity and observed convergence behavior; the sampling interval $k = 10$ balances the amount of training data used and computational efficiency.

\begin{table}[h]
    \caption{Hyperparameters for the simulation\label{tab: Hyperparameters for the simulation}}
    \centering
    \begin{tabular}{cc}
            \hline
            Hyperparameters for the simulation & Symbol \& Value \\
            \hline
            Simulation step size  & $\Delta t = 0.1s$ \\
            Discount factor of value network  & $\gamma  = 0.99$ \\
            Learning rate of adjust temperature & $\lambda  = 3e - 4$ \\
            Learning rate of target network & $\tau_{1}  = 5e - 3$ \\
            Total training steps  & ${n_t} = 100000$ \\
            Sampling interval & $k = 10$ \\
            Same category data size & ${n_{sc}} = 10000$ \\
            Episode number & ${n_{episode}} = 1000$ \\
            Batch size & ${n_b} = 1024$ \\
            Epoch number & ${n_{epoch}} = 1000$  \\
            Lane Width & ${L_{lane}} = 3.5m$ \\
             \hline
            \end{tabular}
\end{table}

\subsection{Simulation Results}

\begin{figure}[htbp]

\centering
\subfloat[]{\includegraphics[width=5.6in]{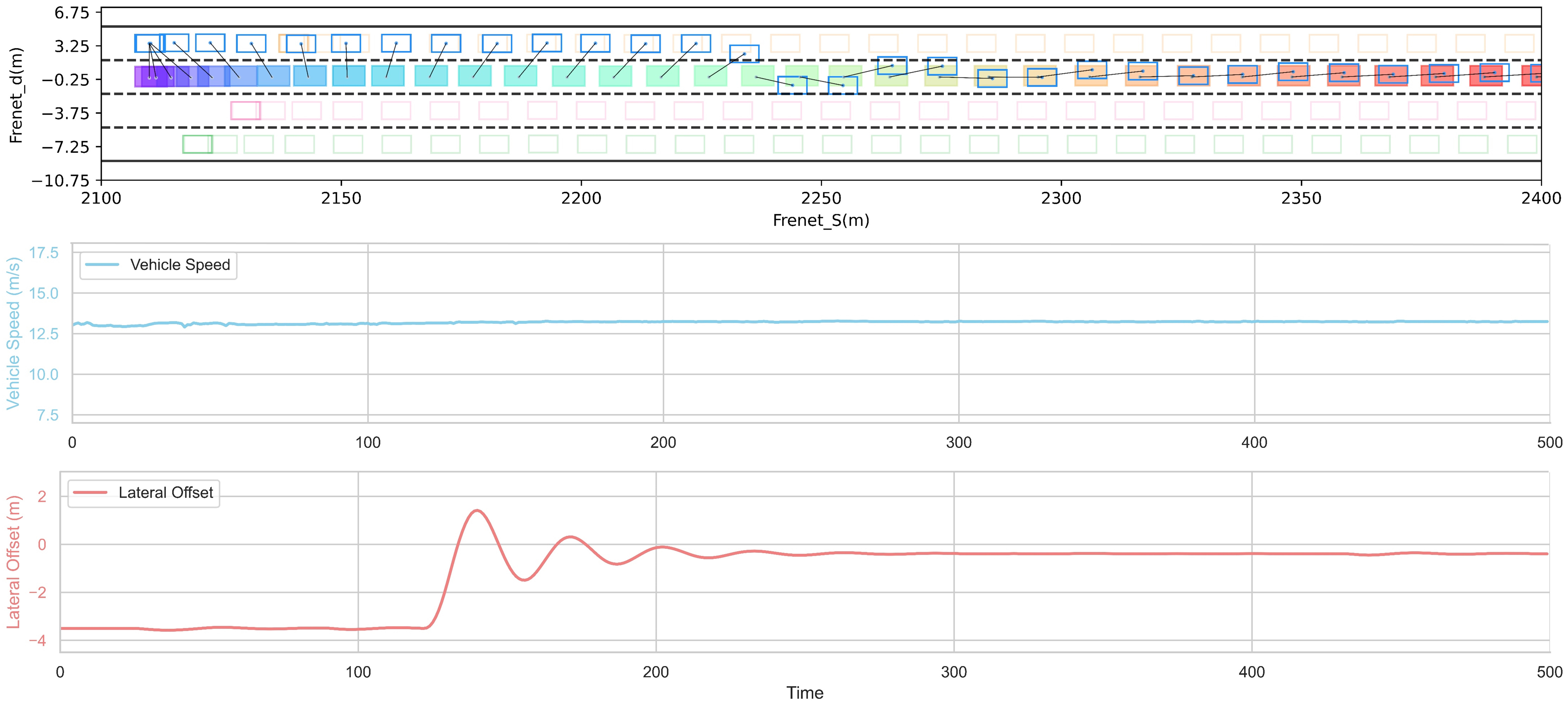}}
\label{adv_scenario_two}
\hspace{0.1in}
\subfloat[]{\includegraphics[width=5.6in]{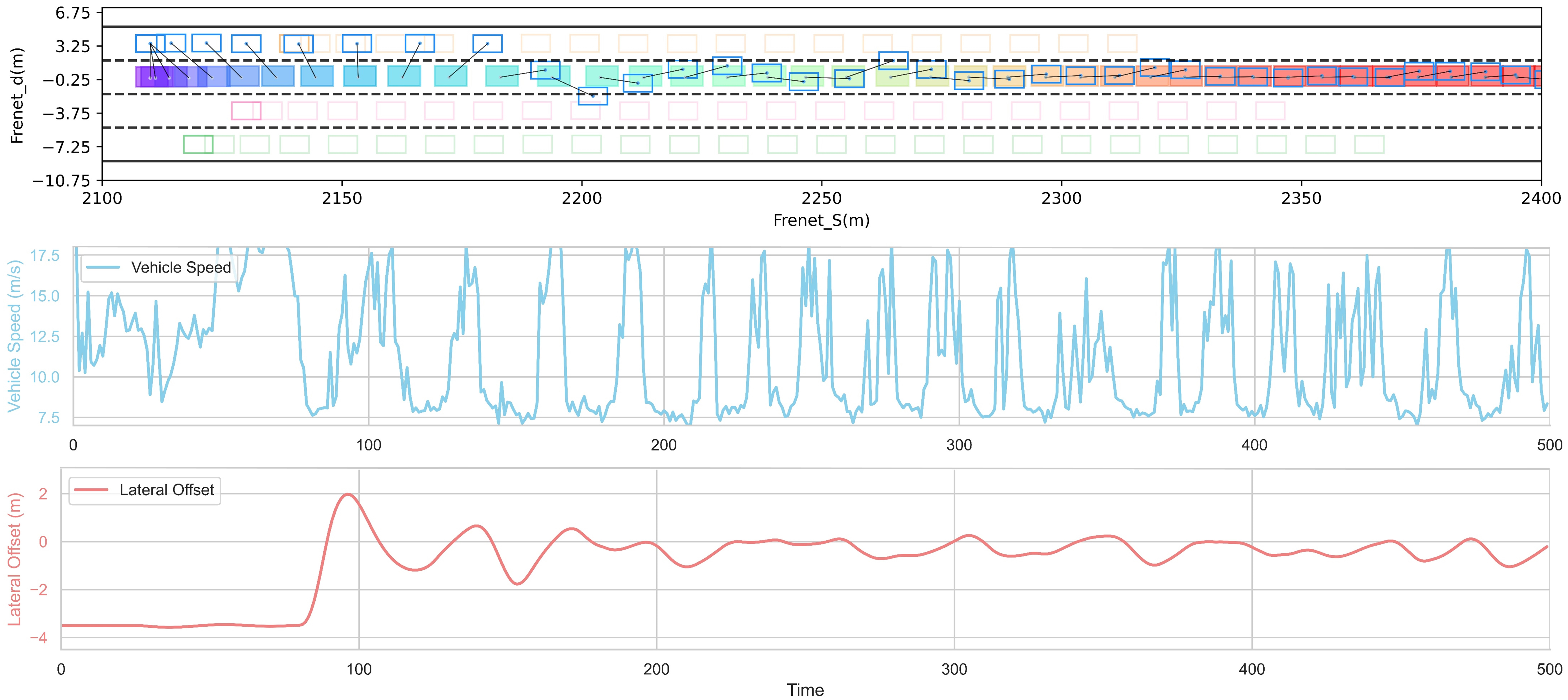}}
\label{adv_scenario_five}
\hspace{0.1in}
\subfloat[]{\includegraphics[width=5.6in]{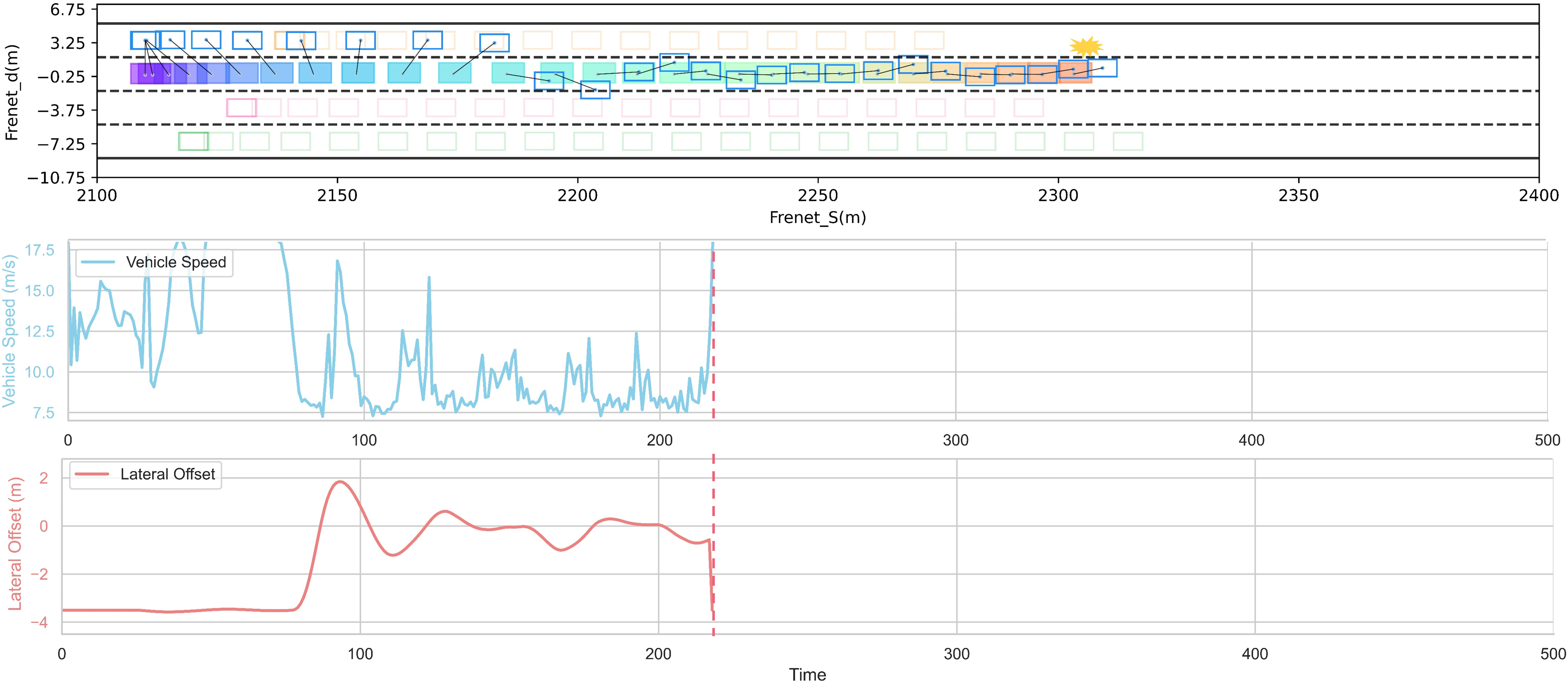}}
\label{adv_scenario_eight}

\caption{The trajectories, velocities and lateral displacements of the ego vehicle and the environment vehicles under various adversarial scenarios of different intensities. (a) Scenario difficulty factor ${X_{scd}} = 0.2$. (b) Scenario difficulty factor ${X_{scd}} = 0.5$. (c) Scenario difficulty factor ${X_{scd}} = 0.8$.}
\label{traj_veloc_lat}
\end{figure}

\begin{figure*}[ht!]
\centering
\begin{tabular}{c}
\includegraphics[width=0.95\textwidth]{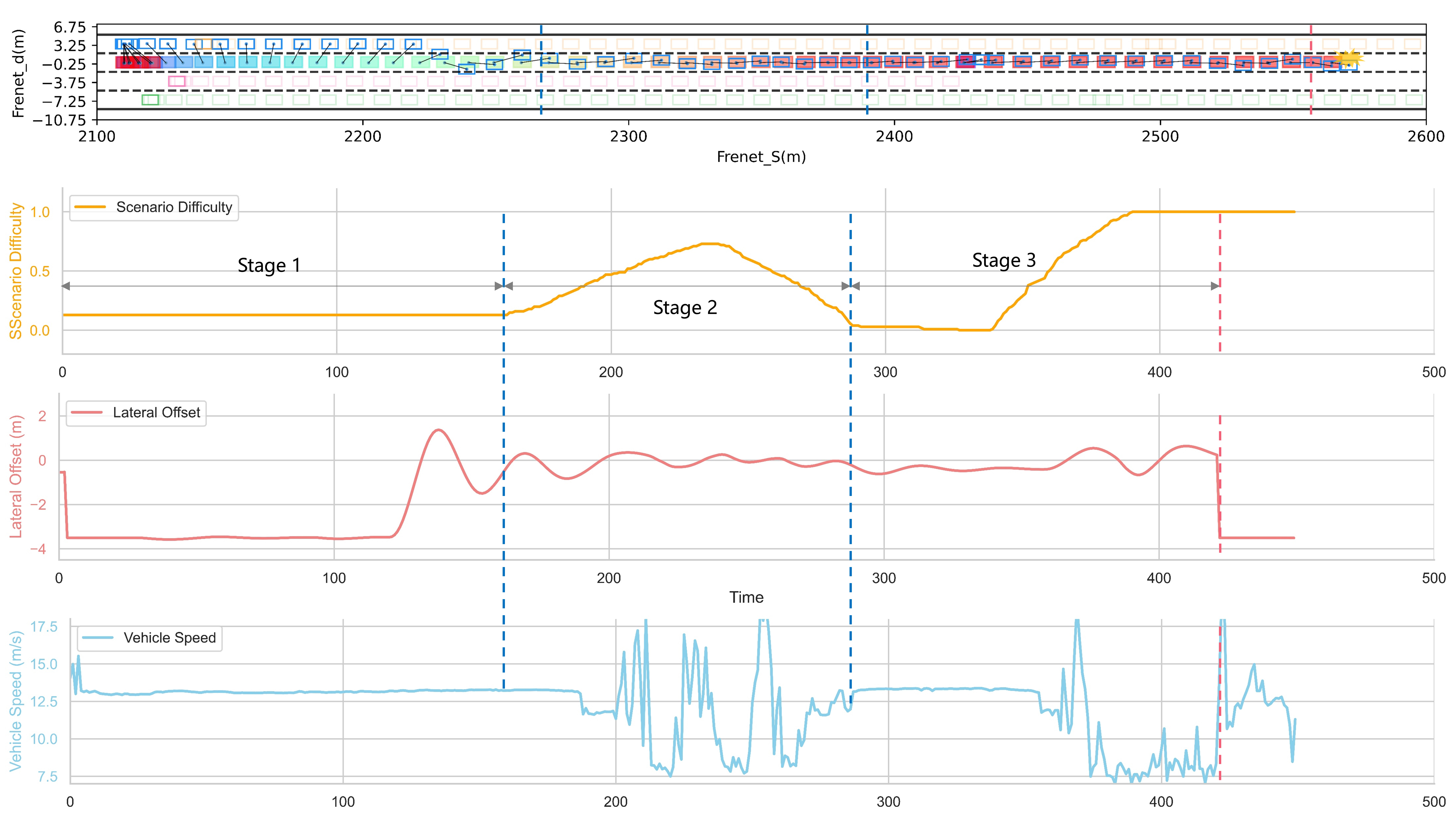}
\end{tabular}
\caption{The trajectory, scenario difficulty, velocity and lateral displacement of the ego vehicle and the environment vehicles under continuously varying scenario difficulty factor}
\label{adv_scenario_diff}
\end{figure*}

\begin{figure}[htbp]

\centering
\subfloat[]{\includegraphics[width=2.9in]{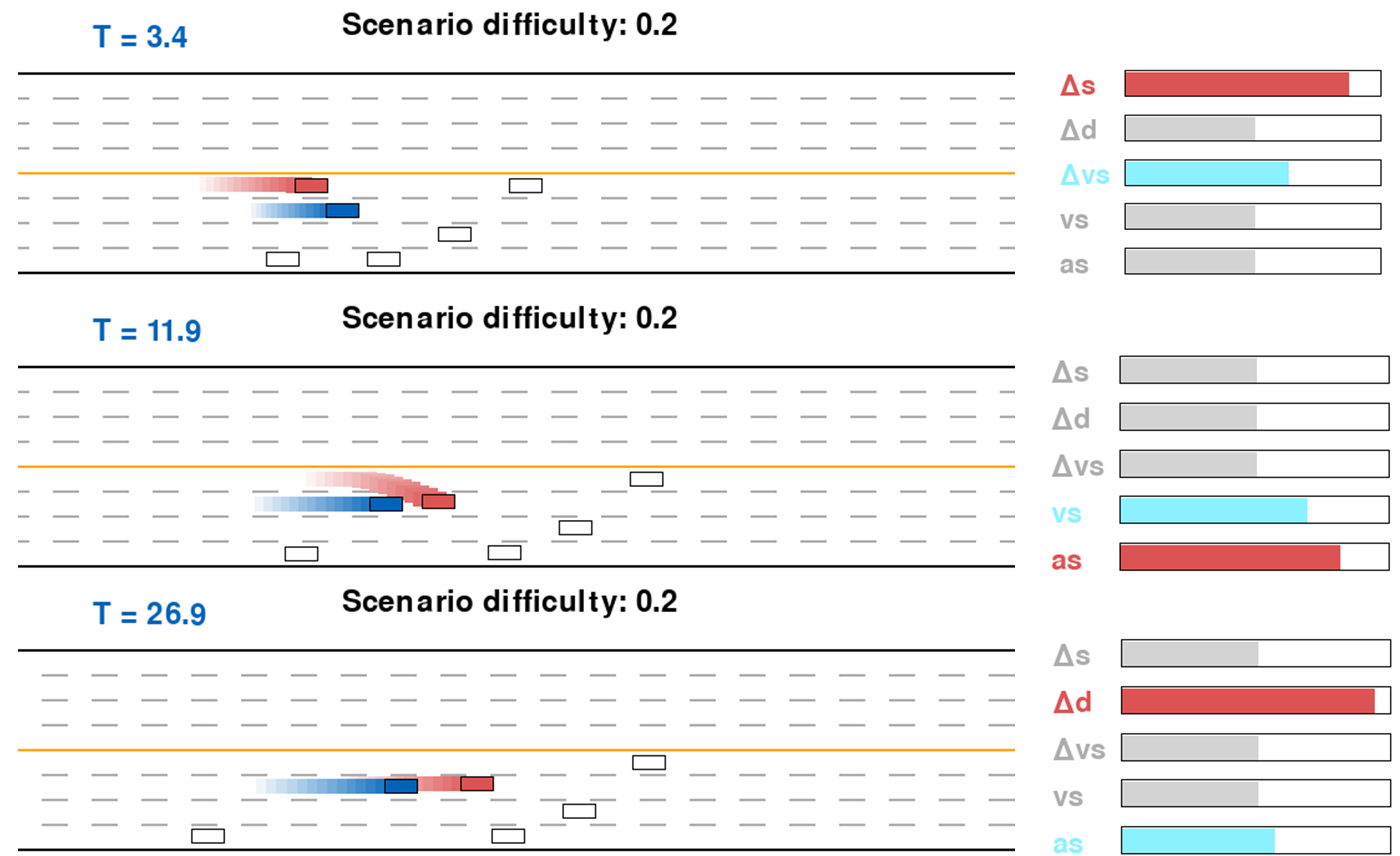}}
\label{QKA_zero_two}
\hspace{0.1in}
\subfloat[]{\includegraphics[width=2.9in]{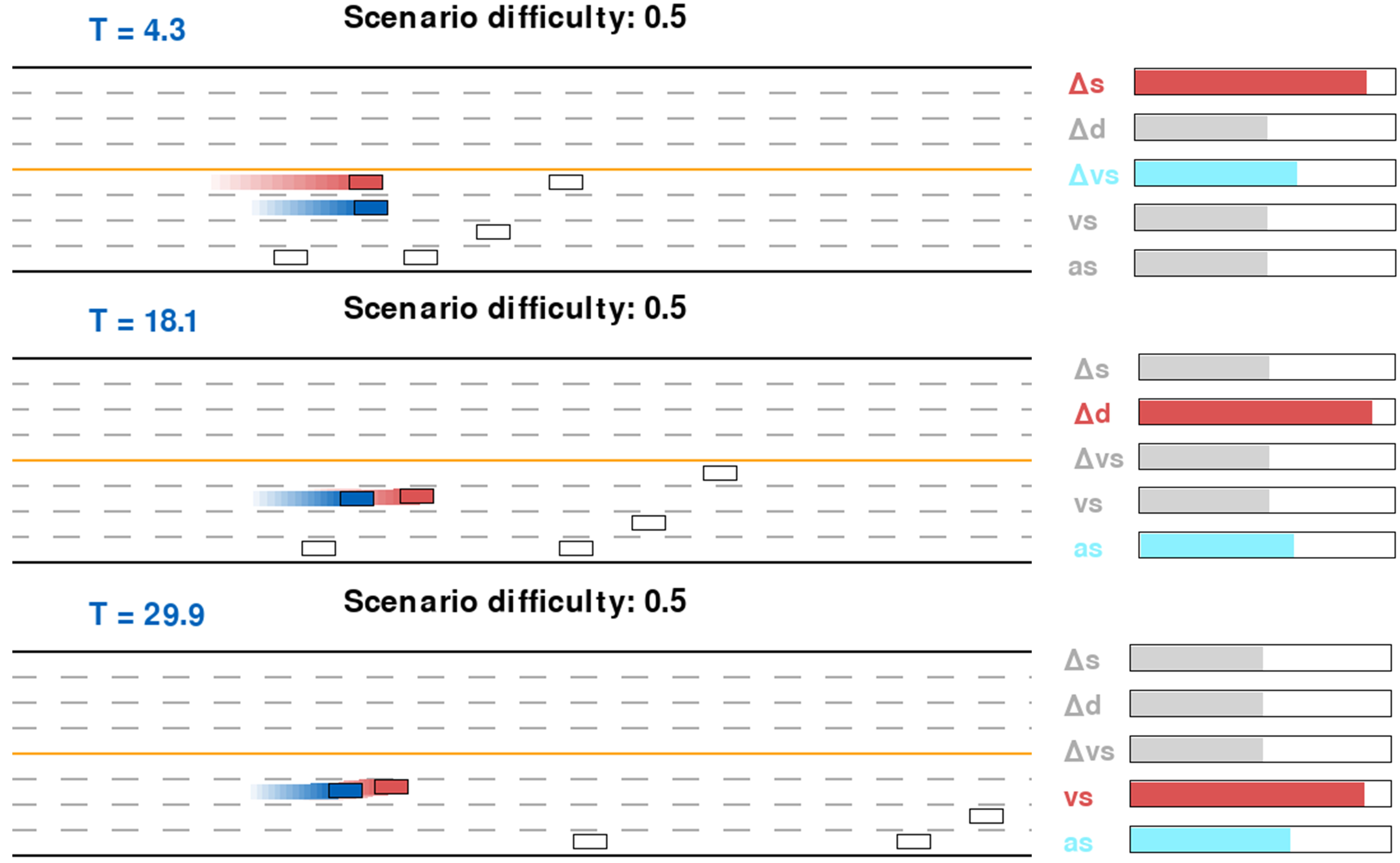}}
\label{QKA_zero_five}
\hspace{0.1in}
\subfloat[]{\includegraphics[width=2.9in]{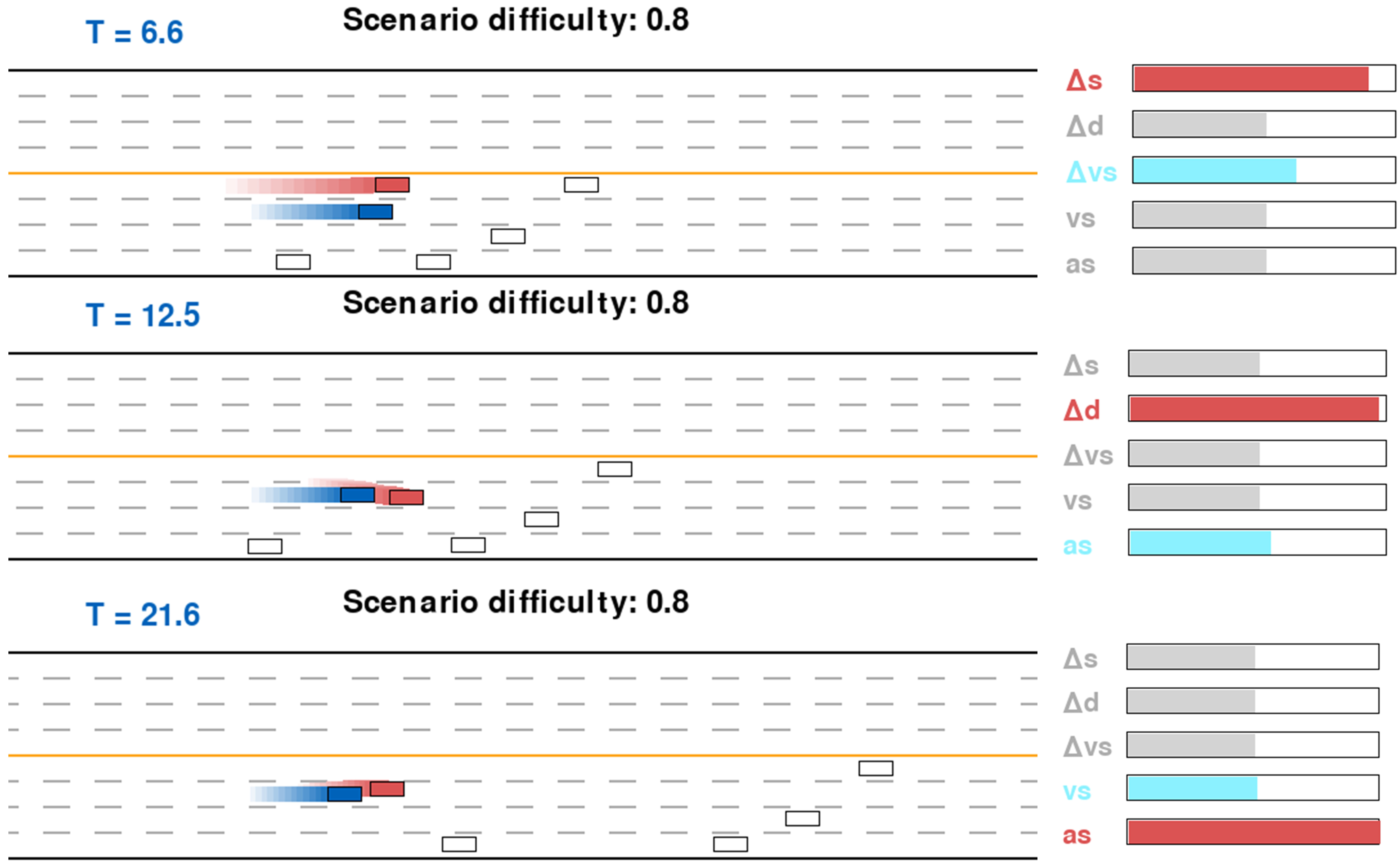}}
\label{QKA_zero_eight}
\hspace{0.1in}
\subfloat[]{\includegraphics[width=2.9in]{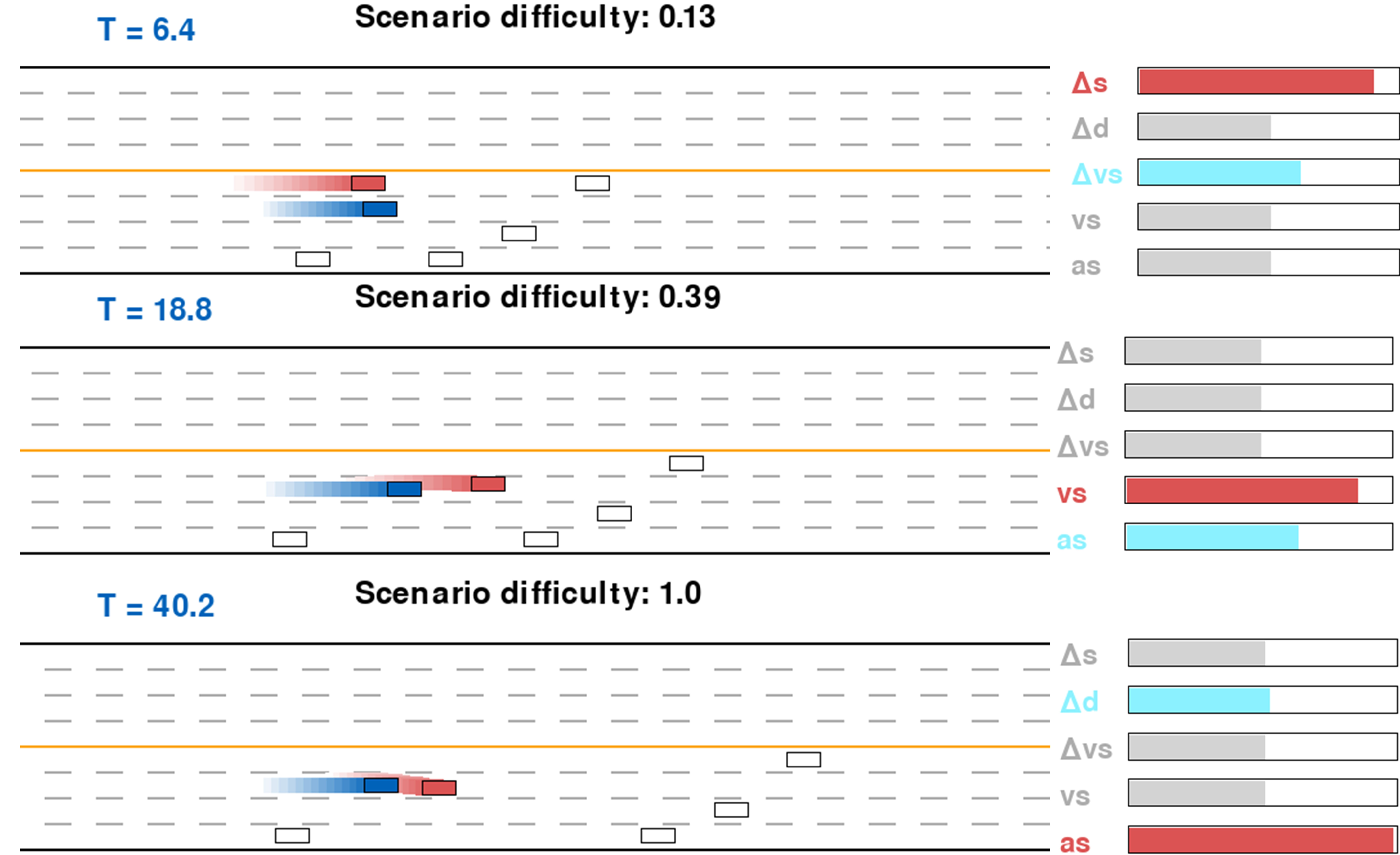}}
\label{QKA_varied}
\hspace{0.1in}
\subfloat[]{\includegraphics[width=2.9in]{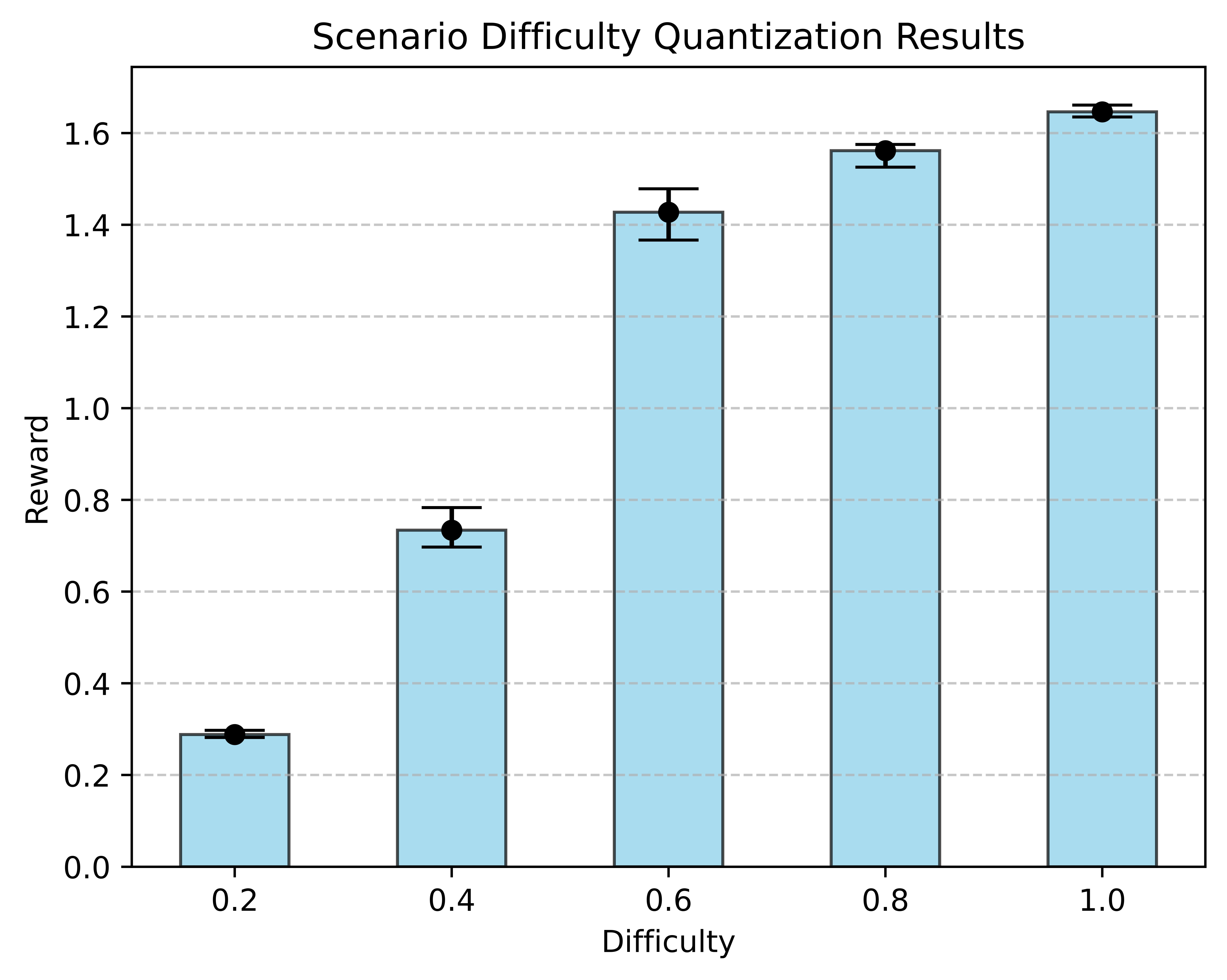}}
\label{reward_bar}
\hspace{0.1in}
\subfloat[]{\includegraphics[width=2.9in]{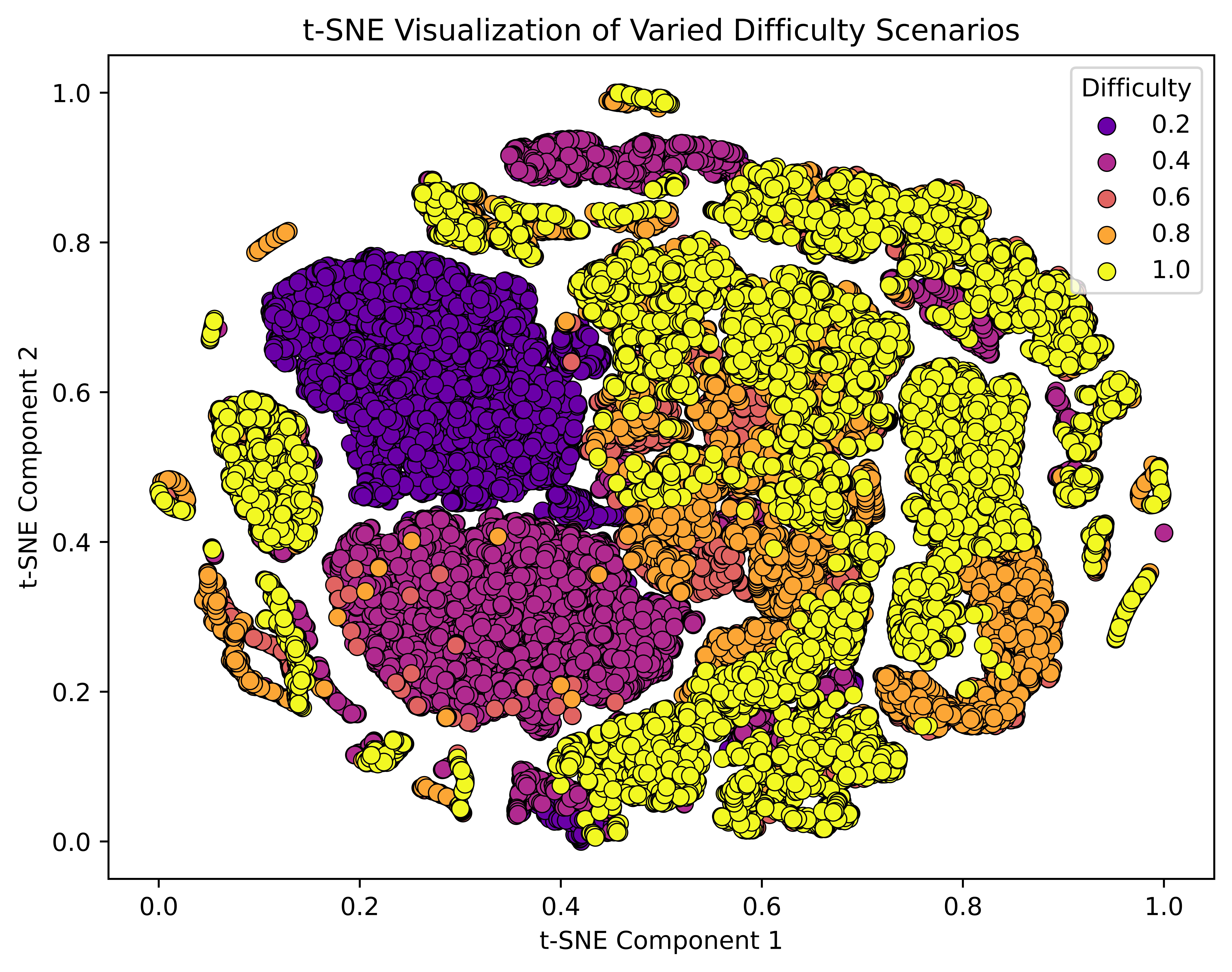}}
\label{TSNE}
`\caption{Visualization results of quantitative representation of scenario difficulty. (a) Scenario difficulty factor ${X_{scd}} = 0.2$. (b) Scenario difficulty factor ${X_{scd}} = 0.5$. (c) Scenario difficulty factor ${X_{scd}} = 0.8$. (d) Scenario difficulty factor ${X_{scd}}$ that dynamically changes with manual input.} (e) Scenario difficulty quantitation results. (f) t-SNE visualization of varied difficulty scenarios.
\label{QKA_TSNE}
\end{figure}

The algorithm is deployed in the training environment. The computer is equipped with an Intel core i7-10700 CPU, NVIDIA GeForce GTX 1660 SUPER GPU. The ego vehicle's policy is chosen as the learning-based autonomous driving algorithm with efficient self-evolutionary capabilities\cite{huang2023efficient}.

\subsubsection{Quantitative result}

The effectiveness of the scenario difficulty generation method proposed in this paper is analyzed through quantitative results.

Fig. \ref{traj_veloc_lat} shows the trajectories, velocities, and lateral displacements of the ego vehicle and the environment vehicles under various adversarial scenarios of different intensities. Specifically, Fig. \ref{traj_veloc_lat}(a)-(c) present the results when the scenario difficulty factor ${X_{scd}}$ is 0.2, 0.5, and 0.8, respectively. When the scenario difficulty factor is low, as in Fig. \ref{traj_veloc_lat}(a), the speed change of the environment agent is smooth, resulting in little effect on the ego vehicle following behind. When the scenario difficulty factor ${X_{scd}}=0.5$, as in Fig. \ref{traj_veloc_lat}(b), the environment agent produces a more pronounced adversarial behavior. This is evident from the increased lateral offset, as it cuts in more aggressively, while also generating sharp acceleration and deceleration to minimize the performance of the ego vehicle. In the most challenging scenario, the environment agent is able to actively resist the ego vehicle by jointly controlling the lateral offset and longitudinal speed until a collision occurs.

Fig .\ref{adv_scenario_diff} shows the trajectory, scenario difficulty, velocity and lateral displacement of the ego vehicle and the environment vehicles under continuously varying scenario difficulty factor. The experiment is divided into three stages. In the first stage (Stage 1), the scenario difficulty ${X_{scd}}=0.13$. At this stage, the behavior of surrounding vehicles is relatively calm, having minimal impact on the ego vehicle, and the vehicle speed is relatively stable, indicating that the behavior generated under low difficulty scenarios is reasonable and smooth. In the second stage (Stage 2), the scenario difficulty gradually increases and then decreases. Surrounding vehicles periodically accelerate and decelerate to confront the ego vehicle, causing fluctuations in vehicle speed. In the third stage (Stage 3), the scenario difficulty is raised to 1, reaching the maximum difficulty. The adversarial behavior of the surrounding vehicles is most intense at this stage, ultimately leading to a collision between the ego vehicle and the surrounding vehicles, verifying the reasonableness of the extreme behavior generated under high difficulty confrontation scenarios.

\subsubsection{Qualitative result}

Through the qualitative results, the reasonability of the proposed method can be further analyzed.

The visualization results of quantitative representation of scenario difficulty is shown in Fig. \ref{QKA_TSNE}.  Fig. \ref{QKA_TSNE}(a)-(c) illustrate the contribution ordering among input state  for the fixed scene difficulty factor (${X_{scd}} = 0.2,0.5,0.8$). It can be seen that as the scenario evolves, the change in the contribution ordering of the state $S$ consists of three phases:

1. Overtaking phase: the environment agent accelerates in order to reach the front of the ego vehicle to ensure that it can have an opportunity to influence the vehicle. At this point state $\Delta s$ has the highest contribution.

2. Cut-in phase: the environment agent in front of the ego vehicle generates adversarial behavior as much as possible through the control of the longitudinal movement. At this point, states ${v_s}$ and ${a_s}$ have the highest contribution.

3. Maintenance phase: the environment agent is in the process of confrontation with the ego vehicle. If there is any lateral offset deviation between the environment agent and the ego vehicle, the environment agent must block the ego vehicle and continue to produce adversarial behaviors. At this stage, the state ${\Delta _d}$ has the highest contribution.

The above analysis process illustrates that the proposed data-driven quantitative representation method of scenario difficulty not only realizes the generation of variable difficulty scenarios, but also has a certain level of interpretability compared with traditional black-box methods.

Fig. \ref{QKA_TSNE}(d) illustrates the scenario difficulty factor ${X_{scd}}$ that dynamically changes with manual input. It can be seen that the proposed data-driven model can generate adversarial scenarios with continuous variable difficulty, and the correlation degree between states is also interpretable.

Fig. \ref{QKA_TSNE}(e) shows the quantization results of the scenario difficulty. The proposed scenario difficulty quantitative representation model is deployed to generate variable difficulty scenarios. The average reward for each episode is saved and used as a quantitative metric. It can be seen that the proposed method can generate diverse scenarios with quantifiable difficulty.

T-SNE (t-distributed stochastic neighbor embedding) is a machine learning algorithm for dimensionality reduction and visualization of high-dimensional data \cite{van2008visualizing}. This method can be used to understand and present patterns and relationships in high-dimensional data. In this paper, the t-SNE method is applied to process the data in the varied difficulty scenario dataset. The t-SNE visualization of varied difficulty scenarios is shown in Fig. \ref{QKA_TSNE}(f). It can be seen that the scenario data of different difficulty levels has obvious aggregation. This shows that the proposed method can generate variable difficulty adversarial scenarios with discrimination.

\section{Conclusion}

This paper proposed a data-driven quantitative representation method of scenario difficulty. The proposed scenario generation model based on the transformer architecture, and it involves two key steps, namely, the policy search of the environment agent and the data generation of scenarios with varying difficulty. The method was validated in a typical adversarial scenario. The experimental results demonstrated that the proposed algorithm can generate reasonable and highly distinguishable scenarios with quantifiable difficulty representations without any expert logic rule design. This applied to both fixed and dynamically changing scenario difficulty factor inputs. Furthermore, the analysis results indicated that the proposed method has a certain level of interpretability compared  to  traditional black-box methods. Future research will include large-scale natural driving datasets to generate more realistic adversarial scenarios. Additionally, the proposed method will be extended to cover more adversarial scenario generation tasks.

\section*{Acknowledgments}

\subsection*{Author Contributions}

S. Yang, and Y. J. Huang conceived the idea and designed the experiments. S. Yang and C. J. Wang wrote the original Draft. Y. J. Zhang, Y. M. Yin, S. E. Li and H. Chen reviewed and revised the draft. H. Chen supervised and led the planning and execution of research activities.

\subsection*{Funding}

We would like to thank the National Key R\&D Program of China under Grant No2022YFB2502900, National Natural Science Foundation of China (Grant Number: U23B2061), the Fundamental Research Funds for the Central Universities of China, Xiaomi Young Talent Program, and thanks the reviewers for the valuable suggestions

\subsection*{Conflicts of Interest}
The authors declare that there is no conflict of interest regarding the publication of this article.

\bibliographystyle{unsrt}
\bibliography{sample}

\end{document}